\documentclass[letterpaper, 10 pt, conference]{ieeeconf}

\IEEEoverridecommandlockouts

\usepackage{graphicx}
\usepackage{siunitx} 
\title{\LARGE \bf
EmbodiedLGR: Integrating Lightweight Graph Representation and Retrieval for Semantic-Spatial Memory in Robotic Agents
}

\author{Anonymous Author(s)}

\author{Paolo Riva$^{1}$, Leonardo Gargani$^{*1}$, Matteo Frosi$^{1}$, Matteo Matteucci$^{1}$%
\thanks{$^{*}$ Corresponding author}
    \thanks{$^{1}$All authors are with the Department of Electronics, Information, and Bioengineering (DEIB),
        Politecnico di Milano, Milan, Italy.
        Emails:
        \tt\small{
            paolo13.riva(at)mail.polimi.it,
            \{leonardo.\allowbreak gargani,\allowbreak
            matteo.\allowbreak frosi,\allowbreak
            matteo.\allowbreak matteucci\allowbreak\}(at)polimi.it
        }
    }%
}

\usepackage{tikz}
\newcommand\submittedtext{%
  \footnotesize © 2026 IEEE.  Personal use of this material is permitted.  Permission from IEEE must be obtained for all other uses, in any current or future media, including reprinting/republishing this material for advertising or promotional purposes, creating new collective works, for resale or redistribution to servers or lists, or reuse of any copyrighted component of this work in other works.}
\newcommand\submittednotice{%
\begin{tikzpicture}[remember picture,overlay]
\node[anchor=south,yshift=15pt] at (current page.south) {\fbox{\parbox{\dimexpr0.95\textwidth-\fboxsep-\fboxrule\relax}{\submittedtext}}};
\end{tikzpicture}%
}

\begin{document}

\maketitle
\thispagestyle{empty}
\pagestyle{plain}

\submittednotice

\begin{abstract}
As the world of agentic artificial intelligence applied to robotics evolves, the need for agents capable of building and retrieving memories and observations efficiently is increasing. Robots operating in complex environments must build memory structures to enable useful human-robot interactions by leveraging the mnemonic representation of the current operating context. People interacting with robots may expect the embodied agent to provide information about locations, events, or objects, which requires the agent to provide precise answers within human-like inference times to be perceived as responsive. While prior work has addressed the memory-building phase, little work has optimized memory representations to enable LLM-based agents to quickly build, traverse, and efficiently retrieve memory entries in real-time scenarios. To address this gap, we propose the Embodied Light Graph Retrieval Agent (EmbodiedLGR-Agent), a visual-language model (VLM)-driven agent architecture that constructs dense and efficient representations of robot operating environments. EmbodiedLGR-Agent directly addresses the need for an efficient memory representation of the environment by providing a hybrid building-retrieval approach built on parameter-efficient VLMs that store low-level information about objects and their observation poses in a semantic graph, while retaining high-level descriptions of the observed scenes with a traditional retrieval-augmented architecture. EmbodiedLGR-Agent is evaluated on the popular NaVQA dataset, achieving state-of-the-art performance in inference and querying times for embodied agents, while retaining competitive accuracy on the global task relative to the current state-of-the-art approaches. Moreover, EmbodiedLGR-Agent was successfully deployed on a physical robot, showing practical utility in real-world contexts through human-robot interaction, while running the visual-language model and the building-retrieval pipeline locally.
\end{abstract}

\section{INTRODUCTION}

\begin{figure}[t]
    \centering
    \includegraphics[width=0.48\textwidth]{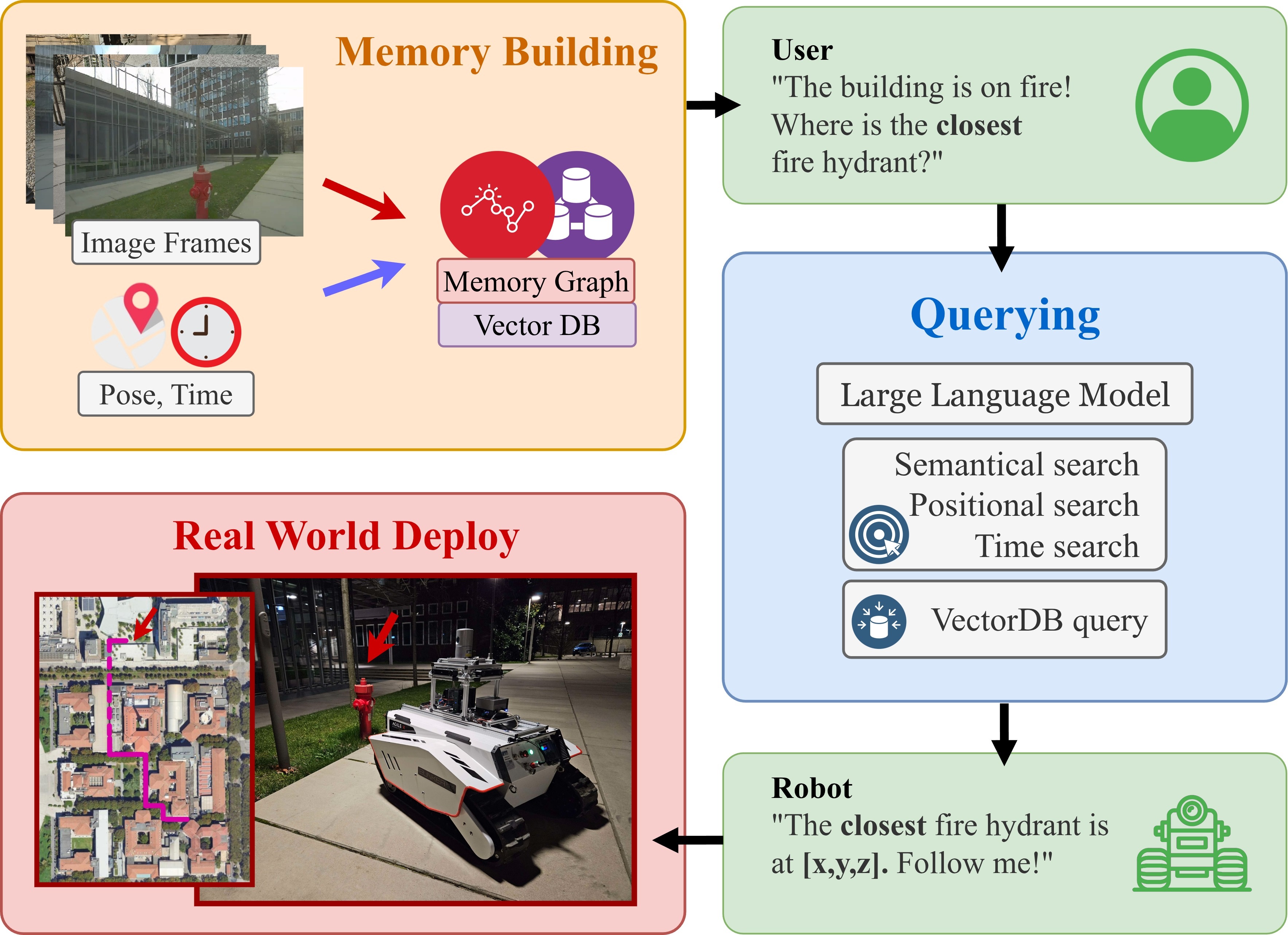}
    \caption{The operational cycle of EmbodiedLGR-Agent starts by (1) first constructing a memory representation of the environment from observations. Then, (2) once sufficient exploration has been conducted,  the user can query the agent with spatial, temporal, or descriptive requests. Lastly, (3) the agent retrieves relevant information from memory and addresses the request, guiding the user towards retrieved positions if needed.}
    \label{fig:intro}
\end{figure}

The deployment of autonomous mobile robots in complex, unstructured environments has driven a growing need for memory-enhanced agentic artificial intelligence. For these systems to engage in meaningful human-robot interaction (HRI)~\cite{zhang2023large} and provide insightful information, they must go beyond executing simple, reactive commands to understand and recall the context of their surroundings, as a human would with past knowledge of a given environment. Humans naturally expect robots operating in domestic or public spaces to answer questions regarding the location of objects, the history of events, or the spatial layout of the environment. To meet these expectations, embodied agents must construct robust memory structures that allow for the rapid and precise information retrieval at human-like inference speeds.

Recent literature has actively explored the integration of Large Language Models (LLMs) and Vision-Language Models (VLMs) to tackle these embodied challenges~\cite{driess2023palm}, where several studies have translated visual observations into structured semantic maps through ad-hoc tailored visual models~\cite{Shafiullah2022Clip, huang2023visual}. At the same time, other works have focused on building expressive 3D scene graphs~\cite{Hughes2022Hydra, Gu2024Conceptgraphs}. Furthermore, to manage long-term context and address complex spatial queries inherent to Embodied Question Answering (EQA) tasks~\cite{Das2018Embodied, Majumdar2024Openeqa}, contemporary approaches increasingly employ Retrieval-Augmented Generation (RAG) paradigms~\cite{Guo2024Lightrag}. Systems such as ReMEmbR~\cite{anwar2025remembr} store episodic observations and VLM-generated captions in vector databases to extend the agent's effective memory. However, many state-of-the-art methods treat memory accumulation monotonically, rely on computationally expensive models, or fail to optimize redundancies and repetitions among semantic concepts. As a consequence, they lead to dense, conceptually overly abundant vector databases that introduce computational overhead, which hinder real-time querying capabilities and spatial reasoning for time-responsive human-robot interactions.

To address this gap, we propose the Embodied Light Graph Retrieval Agent (EmbodiedLGR-Agent), a VLM-driven agent architecture that constructs efficient representations of robot operating environments using lightweight VLMs and memory structures, as illustrated in Fig.~\ref{fig:intro}. EmbodiedLGR directly addresses the need for efficient memory by providing a hybrid building-retrieval approach that stores low-level information about objects and their observation poses within a semantic memory graph, while retaining high-level descriptions of the observed scenes in a separate vector database. This dual-level structure enables the agent to exploit the traversability of graph representations, allowing real-time updates and entity associations over long-horizon image feeds from the robot's exploration.

Our primary contributions are summarized as follows.
\begin{itemize}
    \item We introduce EmbodiedLGR-Agent, a novel architecture that uses a semantic memory graph to handle multiple observations of the same entity/object through real-time positional and temporal updates, while retaining querying capabilities over a vector database, depending on the complexity of the user query. 
    \item We propose a suite of three specialized retrieval tools, i.e., semantic, positional, and time-based, that enable the LLM agent to address diverse queries with low latency.
    \item We demonstrate the efficiency of our approach by evaluating on the NaVQA~\cite{anwar2025remembr} dataset, achieving state-of-the-art inference and querying times for memory items while maintaining competitive accuracy at region-level reasoning and answering.
    \item We validate the practical utility of our system by successfully deploying it on a physical robot and running the entire VLM and memory pipeline locally on the platform during the exploration and querying phases.
\end{itemize}

\section{RELATED WORK}

The development of autonomous robot agents capable of operating in complex, real-world environments has led to a shift from purely traditional utility- and goal-based agents to memory-augmented, context-aware, cognitive architectures. Modern embodied systems must not only perceive their surroundings but also maintain a persistent, long-horizon internal representation of the spatial, temporal, and semantic characteristics of their operational context.

\subsection{Semantic Memory and Scene Representations}

Recent studies have made significant progress in translating 2D visual observations into structured semantic memories for autonomous agents. Some approaches, such as CLIP-fields~\cite{Shafiullah2022Clip} and Visual Language Maps~\cite{huang2023visual}, leverage foundation models and compose them to embed semantic features into spatial maps, enabling text-based retrieval that relies on similarity metrics computed in the proposed representation space. Furthermore, 3D scene graphs~\cite{Hughes2022Hydra, Gu2024Conceptgraphs} have emerged as powerful mental models for agents, providing high-level representations of entities and their spatial relationships by clustering atomic perceptions into hierarchical representations. While these approaches offer semantically rich representations of the environment that aid language-conditioned navigation, they fall short in real-time robotic scenarios, as they do not explicitly optimize the processing pipeline and the data representation for real-world deployment.

\subsection{Retrieval-Augmented Generation for LLMs}

With the rapid advancement of Retrieval-Augmented Generation (RAG) techniques~\cite{Guo2024Lightrag, gupta2024comprehensive}, embodied agents are increasingly relying on external memory structures to circumvent the context-window limitations of traditional LLMs~\cite{glocker2025llm}. In robotics, systems such as ReMEmbR~\cite{anwar2025remembr} have successfully applied RAG by storing episodic observations (e.g., VLM-generated captions) and leveraging an LLM to iteratively query this history to reason over long horizons. While these systems excel at building comprehensive memories, often exploiting computationally-heavy representations to provide several information primitives to the agent~\cite{mao2025meta}, they often struggle with redundancy and fragmentation when tracking dynamic entities, as the memory bank grows monotonically over time and is not optimized to manage semantic redundancy over atomic pieces of information.

\subsection{Embodied Question Answering}

Embodied Question Answering (EQA)~\cite{Das2018Embodied} activities require agents to organically integrate perception, navigation, and reasoning to answer natural language queries about their environment. Recent large-scale benchmarks, such as OpenEQA~\cite{Majumdar2024Openeqa}, challenge agents to perform spatial reasoning similar to human cognitive processes, demanding a deep understanding of what the robot has seen, as well as where and when it was observed. While prior works rely on generic querying against monolithic memory banks, achieving human-like response times and low-latency, real-time processing of scenes with memory-efficient visual language models like \textit{Florence-2}~\cite{xiao2024florence} remains mostly unexplored.

\section{METHOD}

\begin{figure*}[t]
    \centering
    \includegraphics[width=0.99\textwidth]{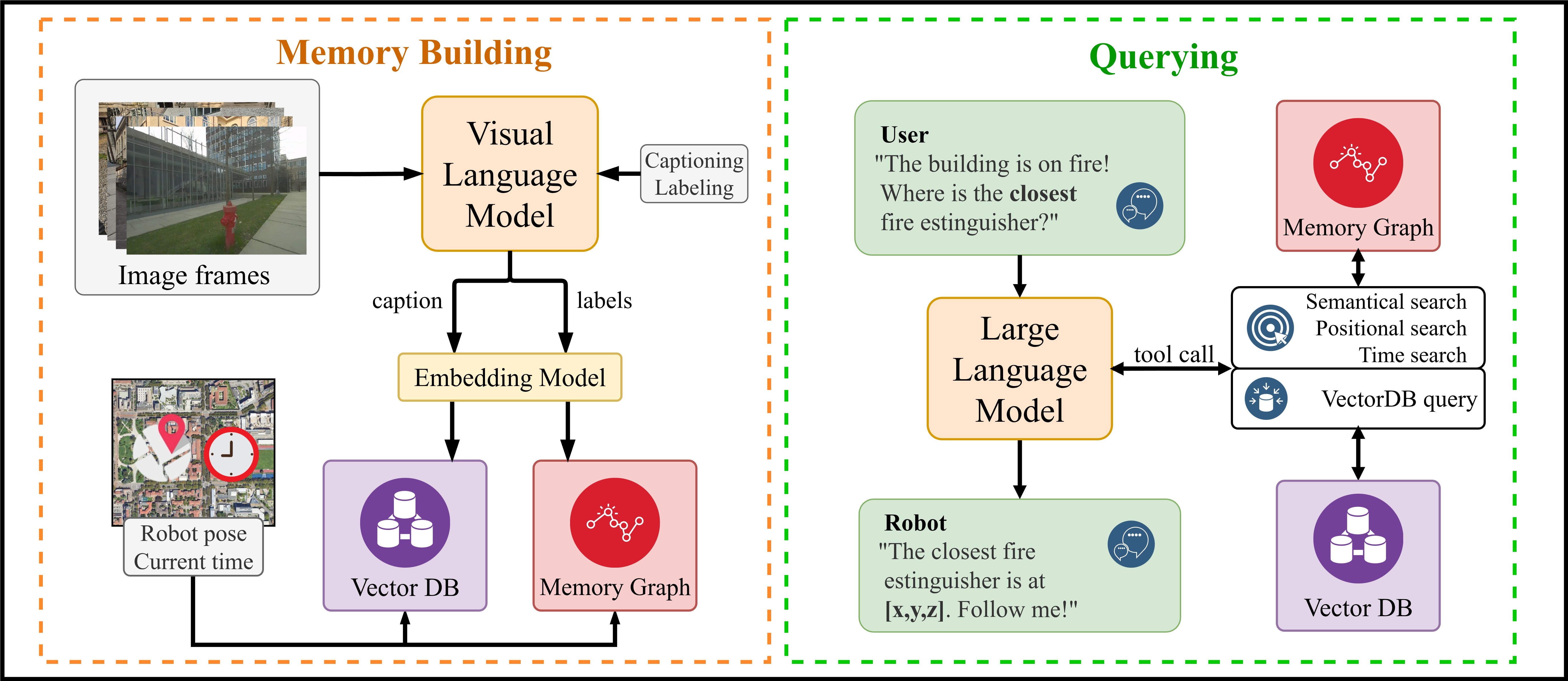}
    \caption{We split the EmbodiedLGR-Agent operational phases between memory building (left) and querying (right). The memory building phase ingests frames from the robot's image perception suite (i.e., one or more RGB cameras) and feeds them to a \textit{Florence-2} VLM, generating, for each frame, a caption and a set of object labels. These are then used to populate the vector database and the memory graph, along with the robot's pose and time. The querying phase begins when the user submits a query, after which an LLM is invoked to retrieve memory data via provided tools from either the memory graph or the vector database. The LLM provides an answer to the user based on what it retrieved during the reasoning loop.}
    \label{fig:agent_overview}
\end{figure*}

EmbodiedLGR-Agent proposes a VLM-driven memory architecture for LLM-based agents to populate and manage memory elements from the robot's observations, adopting a double-level approach to enable effective and efficient information retrieval in real-world robotic contexts. Therefore, our work focuses on providing a memory representation that is efficient in both the memory building phase and the retrieval phase, while retaining the flexibility to query across multiple information dimensions, from positional to semantic. An overview of the agent's architecture is shown in Fig.~\ref{fig:agent_overview}. 

EmbodiedLGR-Agent directly addresses the need for an efficient memory representation of the environment by introducing three retrieval tools for agents and providing two logically separated memory structures: a memory graph and a vector database. Through such tools, the LLM agent can address user queries directly from the memory graph, either performing semantic-based retrieval, positional-based retrieval, or time-based retrieval. The vector database, following state-of-the-art approaches, is populated with richer semantic descriptions of the observed scenes and queried only by the agent for semantically complex user queries.

\subsection{Memory Building}
\label{sec:mem_build}

The memory building phase begins (1) with the collection of the current image frame, position and timestamp from the exploring robot; (2) within the same time instant, the frame is processed by a VLM to extract all of the objects present in the scene, along with the visual description of the scene itself; (3) once the set of object labels and frame description are returned, embeddings for each label and the description are generated, to (4) populate the memory graph and the vector database. Multiple perceptions of the same object within the current observation window are updated on the memory graph, avoiding subsequent duplicate entries.
Moreover, we define the entity data structure as a spatial-semantic proximity graph, as nodes are inserted and linked directly by spatial proximity and embedding space distance, allowing the retrieval step of the agent to expand in the agent's context window only the top-$k$ set of relevant nodes without the need to explore the full memory graph for a given entity query. 
Indeed, the graph represents nodes in both the spatial and text embedding reference systems, allowing for the definition of query-specific tools presented in the Memory Retrieval Section.

\textbf{Data Processing.} Our proposed method considers a stream of image frames $I_{1:k}$, a set of poses $P_{1:k}$ and a set of time instants $T_{1:k}$. We associate each image $I_i$ with a unique pose $P_i$ and time $T_i$, corresponding to the robot's pose and time at which the image was sampled from perception. Image frames, poses, and time instants are therefore sampled from the robot as it moves through the environment.
Each image frame $I_i$ is processed in real-time by a visual-language model, extracting both a set of object labels $\{O_1, \dots, O_n\}_i$, corresponding to brief semantic descriptions of the detected objects, and a global visual description $D_i$ of the frame. 

Both outputs are obtained by running \textit{Florence-2} locally on the robot, a multi-purpose, lightweight VLM that processes images with low inference times and performs both object detection and frame captioning.
Labels $O_{1:n}$ and description $D_i$ associated to each image frame $I_i$ are then processed by the sentence transformer \textit{all-MiniLM-L6-v2}~\cite{wang2020minilm} to compute label embeddings $\{E_1, \dots, E_n\}_i$ and description embedding $\mathbf{e}_i$ respectively.

\textbf{Memory Graph Population and Update.} Given an image frame $I_i$ with $n$ objects being observed, triplets of the form $\{E_{j}, P_i, T_i\}$ are created, with $j\in{1:n}$, associating entities in the frame to the robot's pose and time when they were detected during exploration. From here, each triplet is stored as a node in the memory graph for future retrieval. The implemented memory graph is therefore populated with low-level semantic information about observed objects and is directly accessible to the robot's agent at any time. 

The traversing efficiency of the memory graph allows the system not only to create nodes from novel observations but also to update nodes in real time that correspond to previously observed objects or entities. Unlike other state-of-the-art approaches~\cite{anwar2025remembr}\cite{mao2025meta}, which base their work only on heavy and data-wise redundant representations of the scenes, we exploit the traversability and query-responsiveness of the memory graph representation to handle multiple observations of the same entity in subsequent frames, by updating the pose and time of detection on already-created node instances. 
Consider an already-observed object represented by the node $\{E_j, P_h, T_h\}$, with embedded label $E_j$, last-associated pose $P_h$, and last-observed at time $T_h$. Upon observing the same embedded label $E_j$ in a new pose $P_k$, and time instant $T_k$, the node's associated pose is updated if:
$$
||P_k - P_h|| \leq \delta_P,\eqno{(1)}
$$
where $\delta_P$ is a distance threshold specified as a parameter of the agent, depending on its context of deployment. Specifically, if the condition is satisfied, the pose is updated on the components $\{x,y,z\}$ with a moving-average approach between $P_h$ and $P_k$, while the pose's original yaw is preserved for simplicity, and the node is updated with the new time $T_k$. If the condition over $\delta_P$ is not satisfied, a new node with embedding $E_j$, pose $P_k$, and time $T_k$ is created.

Since the VLM can attribute multiple semantically-equivalent labels for the same entity between subsequent image frames (e.g., ``cup" and ``mug" retain very similar semantic meaning, therefore being close in the embedding space), we also consider a semantic similarity threshold $\delta_E$ to avoid duplicate creation of the same entity during detection, following the cosine similarity formula:

$$
\frac{E_h \cdot E_k}{||E_h||_2 ||E_k||_2} > \delta_E, \eqno{(2)}
$$
where $E_h$ and $E_k$ are two label embeddings being observed in subsequent frames associated with positions within distance $\delta_P$.
If the cosine similarity score between the two labels is above the defined $\delta_E$, the already-existing instance of the object with label embedding $E_h$ is updated as previously described, as the pipeline considers the new observation embedded as $E_k$ to match an existing node $E_h$ semantically.

EmbodiedLGR-Agent also handles multiple objects of the same type being observed in a single image frame $I_i$, by checking the number of semantically-equivalent nodes already present in memory and comparing it with the number of currently observed equal entities. Suppose we observe $k$ elements yielding the same embedding $E_j$. Given the considered distance radius $\delta_P$, assume the memory graph already retains $h$ elements with label $E_j$. From here, based on $k$ and $h$, our approach proceeds by (1) updating the position and time on nodes associated with the previously observed $h$ entities, or (2) creating new entity nodes if the current observed scene contains more elements of the considered label than the number of already-known ones, i.e., $k-h$. For scenarios where the observed entities are fewer than the ones already on the graph, i.e., $k<h$, we update the $k$-closest nodes with respect to the current position. This approach, with sufficient exploration of the environment, allows the discovery and update of all objects of the same type $E_j$ within a positional radius $\delta_P$.

\textbf{Vector Database Population.} As the robot populates the memory graph, it simultaneously fills the vector database with a richer representation of the observed scene frame, by generating a visual description from the VLM and embedding it through the \textit{all-MiniLM-L6-v2} model as described in the Data Processing phase. Triplets of the form $\{\mathbf{e}_i, P_i, T_i\}$ are built and inserted into the vector database, associating description embedding $\mathbf{e}_i$ with position $P_i$ and time $T_i$.

\subsection{Memory Retrieval}
\label{sec:mem_querying}

Having populated both the memory graph and the vector database, the LLM-agent's querying phase is modeled as a cyclic finite-state diagram, in which the agent invokes retrieval tools based on the user query until an answer is available or the cycle iteration limit is reached. 

EmbodiedLGR-Agent defines three on-graph search tools (callable functions) that retrieve data from the memory graph structure: semantic, positional, and temporal search. Depending on the complexity and nature of the user query, the LLM agent can also invoke querying tools on the vector database by calling a ReMEmbR agent~\cite{anwar2025remembr}, providing articulated answers by reasoning over the stored descriptive captions, as previously described. 
Defining specific tools for each querying approach allows traversing the memory graph from a given primitive (semantics, position, or time) and addressing the request from the retrieved nodes: for example, a request on the position of an object could be addressed by searching for the object's label semantically and returning the associated position as an answer to the user.
Any output returned by the invoked tools is directly added to the LLM context window.
A general outline of the Memory Retrieval process is provided in the Querying section of Figure~\ref{fig:agent_overview}.

\textbf{On-graph Semantic Search.} 
The agent can retrieve nodes that are semantically similar to a query chosen upon tool invocation. The semantic search tool $t_S$ is defined as:
$$
t_S({query}) \rightarrow \{E,P,T\}_{1:k},\eqno{(3)}
$$
where $query$ is provided by the agent, and $k$ nodes are returned to the agent from the graph. The tool first invokes \textit{all-MiniLM-L6-v2} to compute the embedding of the input query, i.e., $Q_E$. Then, it computes the cosine similarity metric over the node-label embeddings, returning the $k$ nodes with embedding $E$ closest to $Q_E$ in the embedding space.

\textbf{On-graph Positional Search.} The agent can retrieve nodes corresponding to objects spatially close to a given position, provided as input by the agent upon tool invocation. The positional search tool $t_P$ is defined as:
$$
t_P({x,y,z}) \rightarrow \{E,P,T\}_{1:k},\eqno{(4)}
$$
where position $\{x,y,z\}$ is provided as input by the agent, and $k$ nodes are retrieved from the graph. The tool computes the L2 distances from the provided position and returns the $k$ nodes with the smallest distances.

\textbf{On-graph Temporal Search.}
The agent can retrieve nodes based on a reference time instant provided as input by the agent. The temporal search tool $t_T$ is defined as follows:

$$
t_T({hh,mm,ss}) \rightarrow \{E,P,T\}_{1:k},\eqno{(5)}
$$
as the timestamp $\{hh,mm,ss\}$ is provided by the agent, and $k$ nodes are returned from the graph having $T$ closest to the input timestamp with respect to the L1 temporal distance.

\textbf{Vector Database Queries.} To query the vector database, EmbodiedLGR-Agent integrates and invokes tools specified by a ReMEmbR agent~\cite{anwar2025remembr} instance, to retrieve either descriptive, positional, or temporal information over the observed scene's descriptions, populated during the memory building phase. The vector database retains visual descriptions of the perceived scenes during the exploration phase. Nonetheless, the LLM agent is prompted to independently decide whether to invoke a retrieval from the vector database, and is prompted to do so only if the primitives available on the memory graph are insufficient for the current query.

Our layered approach allows the EmbodiedLGR-Agent to access only the memory structure that best suits a given query's information request, maximizing retrieval efficiency: for simple, low-level queries (e.g., known position of atomic objects), the agent can perform retrieval on the memory graph through the tools $t_S, t_P, t_T$, while for semantically-complex queries (e.g., observed scene matching a specific scene setting) the agent can perform retrieval directly on the vector database through the ReMEmbR agent. This not only improves the responsiveness of the proposed EmbodiedLGR-Agent over simple, atomic queries but also reduces overhead compared to other presented solutions~\cite{anwar2025remembr}\cite{mao2025meta}.

\subsection{Inference}

The inference process of the EmbodiedLGR-Agent follows the traditional LLM invocation in agentic applications, exploiting system prompting and state-of-the-art libraries to implement agentic components and perform API calls, depending on the chosen model, in an extensible manner.

After receiving a user query via human-robot interaction, the LLM enters a reasoning loop, invoking memory retrieval tools based on the user request's complexity and atomic requests and selecting them independently without strict system prompt constraints. All the information retrieved by the tools over either the graph memory or the vector database is fed into the LLM's context window; tools can be invoked multiple times during the reasoning loop, as illustrated in Fig.~\ref{fig:agent_overview}, until the LLM can provide an answer to the user.

\begin{table*}[htbp]
    \centering
    \caption{Performance comparison of different VLMs and configurations over query accuracy.}
    \label{tab:vlm_results_accuracy}
    \begin{tabular}{llccc}
    \hline
    \textbf{VLM} & \textbf{Agent Configuration} & \textbf{Positional Accuracy [\%]} & \textbf{Temporal Accuracy [\%]} & \textbf{Average [\%]} \\
    \hline
     & Graph Memory & 13.75 & 20.48 & 17.11 \\
    Florence-2-base & ReMEmbR & 6.67 & \textbf{38.63} & 22.65 \\
     & Graph Memory + ReMEmbR & \textbf{19.59} & 30.30 & \textbf{24.94} \\
    \hline
     & Graph Memory & 13.75 & 14.82 & 14.29 \\
    Florence-2-large & ReMEmbR & 19.59 & \textbf{38.63} & 29.11 \\
     & Graph Memory + ReMEmbR & \textbf{29.17} & \textbf{38.63} & \textbf{33.90} \\
    \hline
    \end{tabular}
\end{table*}

\begin{table*}[htbp]
    \centering
    \caption{Performance comparison of different VLMs and configurations over response latency and fallback percentage.}
    \label{tab:vlm_results_latency}
    \begin{tabular}{llcc}
    \hline
    \textbf{VLM} & \textbf{Agent Configuration} & \textbf{Response Latency [s]} & \textbf{Fallback Percentage [\%]} \\
    \hline
     & Graph Memory & 10.80 & --- \\
    Florence-2-base & ReMEmbR & 19.14 & --- \\
     & Graph Memory + ReMEmbR & 23.13 & 93.33 \\
    \hline
     & Graph Memory & 9.97 & --- \\
    Florence-2-large & ReMEmbR & 19.79 & --- \\
     & Graph Memory + ReMEmbR & 23.73 & 80.83 \\
    \hline
    \end{tabular}
\end{table*}

\section{EXPERIMENTAL SETUP}
\label{sec:experimental_setup}

We evaluate the EmbodiedLGR-Agent on the NaVQA dataset~\cite{anwar2025remembr} to assess the agent's ability to build and retrieve memory for navigation-related tasks, which require both semantic understanding of the query and positional retrieval of the target position. We evaluate the responsiveness of the memory graph and the vector database presented, and report the overall accuracy across spatial and temporal metrics. We test object labels and visual captions generation with \textit{Florence-2-base} (0.22B parameters) and \textit{Florence-2-large} (0.77B parameters), two models significantly smaller than the ones employed by other solutions, like \textit{VILA1.5-13B} (13B parameters)~\cite{anwar2025remembr} or \textit{Qwen2.5-VL-7B} (7B parameters)~\cite{mao2025meta}, to further stress the requirement of a feasible deployment scenario for a real-time constrained computational stack.
Label and caption generation are handled by specific tasks available on the chosen VLM. \textit{Florence-2} is a multi-purpose model that allows performing several visual processing tasks on an input image. Specifically, we select the \textit{MORE-DETAILED-CAPTION} task to generate visual captions over frames (populating the vector database), and the \textit{DENSE-REGION-CAPTION} task to generate semantic labels of the entities within the frames (populating the memory graph).

\textbf{Metrics for Evaluation.} Considering the NaVQA dataset and following the approach of ReMEmbR~\cite{anwar2025remembr}, we define specific evaluation criteria to measure the system's ability to answer spatial and temporal questions efficiently.
\begin{itemize}
    \item \textit{Response Latency:} queries are evaluated over the memory graph and vector database depending on the quanta of information to retrieve to return an answer. We evaluate the average response latency across queries in NaVQA and report the latency for each component of the agent's memory architecture.
    \item \textit{Fallback percentage:} the LLM determines which memory structure to query based on the semantic complexity of the request. To show how frequently the agent falls back from just querying the graph to passing the query over the vector database, we consider the number of ReMEmbR agent calls $N_{vector}$ over the number of queries $N_{queries}$ as:

    $$
    Fallback = \frac{N_{vector}}{N_{queries}}.\eqno{(6)}
    $$
    \item \textit{Spatial Accuracy:} given queries requiring the agent to provide the position of an object or an area within the environment, the agent is evaluated over the mean Euclidean distance (L2 norm) between the predicted and ground-truth $\{x,y,z\}$ coordinates. We consider a spatial query to be correct if the error is within 25 meters of the ground truth, as this radius effectively captures region-level retrieval in a real-time deployment with reduced frame rate processing, conditioning the amount of spatial information processed.
    \item \textit{Temporal Accuracy:} given queries requiring the agent to provide the observation time instant of a given object or area, the agent is evaluated over the mean temporal error (L1 norm) with respect to the ground truth. We consider a temporal query to be correct if the error is within 3 minutes of the ground truth, given the coarse temporal sampling rate of a real-time scenario, consistent with the expected performance by the VLM.
\end{itemize}
The adopted error metrics for spatial and temporal accuracy are consistent with the mean and variance of the errors reported in ReMEmbR~\cite{anwar2025remembr}, and they are chosen to evaluate EmbodiedLGR on region-level reasoning. Indeed, Florence-2 is a very compact family of models with respect to the VLMs tested by ReMEmbR, constraining the quality of semantic inference in favor of an edge-deployable VLM solution.

\textbf{Implementation Details.} During the memory-building phase, the agent processes incoming frames at a sub-sampled rate of 1 frame every 2.0 seconds. This extraction rate was empirically chosen to balance high-fidelity spatial tracking with memory efficiency: in real-world deployment scenarios, reduced processing frame rates represent a good compromise between computational overhead and information redundancy. The VLM generation process is limited to 1024 tokens, which we found to be a sufficient upper bound for describing highly cluttered environments without degrading throughput. We also empirically set a distance threshold $\delta_P$ to 5 meters and a semantic threshold $\delta_E$ to 0.75, which we observed to be the best-performing setup for the node-update dynamics described in Section~\ref{sec:mem_build}. As previously specified, we select the \textit{all-MiniLM-L6-v2} sentence transformer to compute embeddings across the architecture.

For agent deployment, we use \textit{GPT-4o} as the reference LLM, invoking it via its APIs alongside our tools through LangChain~\cite{hurst2024gpt}. 
The LLM remains the only cloud component of the system, as local implementations of smaller, tool-augmented language models have been shown to significantly degrade overall performance~\cite {anwar2025remembr}.
The memory graph and the vector database are implemented using NetworkX~\cite{hagberg2007exploring} and Milvus~\cite{wang2021milvus} respectively, enabling a scalable physical deployment of the architecture.

\section{Results}

We evaluated a set of 120 questions from the NaVQA dataset, using both \textit{Florence-2-base} and \textit{Florence-2-large} in combination with either the graph memory, the vector database (queried with a ReMEmbR agent), or both, to assess performance under the ablation of single memory components. This was carried out by selectively passing only a subset of the tools defined in Section~\ref{sec:mem_querying} to the agent.

\textbf{Query Accuracy.} Table~\ref{tab:vlm_results_accuracy} presents the performance comparison over the positional accuracy and temporal accuracy metrics over different versions of the \textit{Florence-2} VLM. The table first shows that combining the vector database (ReMEmbR) with the graph memory component improved the average accuracy relative to the ReMEmbR baseline on the considered query data. By integrating querying capabilities over graph memory along with the vector database (``Graph Memory + ReMEmbR" in the table), the EmbodiedLGR-Agent improved its positional accuracy observed across the dataset. This is likely due to a more fine-grained positional representation being available to the agent, which distributes atomic poses across the semantic entities in the memory graph. For temporal queries, the agent matched ReMEmbR's vector database-only performance for \textit{Florence-2-large}, while showing a bigger difference in performance for the graph-only configuration, as temporal requests from NaVQA were not generally tied to a specific entity in the space, and could be better addressed by querying over the visual captions in the vector database. 
Indeed, for \textit{Florence-2-large}, Table~\ref{tab:vlm_results_accuracy} underlines how positional queries, often linked to a specific entity placement in the space being requested, benefit from the graph memory component retaining entity-pose primitives; temporal queries, on the other hand, are better addressed by querying the vector database, as the combined approach scores an equivalent result without the entity graph. This is also coherent with the Fallback Percentage shown in Table~\ref{tab:vlm_results_latency}.

The results shown in Table~\ref{tab:vlm_results_accuracy} are consistent with the VLM dimensions (parameter size $<1B$), employed for generating visual descriptions and entity labels, and they show that agents like EmbodiedLGR-Agent, which rely on such VLMs, can be effectively deployed for edge-memory building with competitive performance over region-level query accuracy.

\textbf{Latency and Fallback Percentage.} Table~\ref{tab:vlm_results_latency} first demonstrates how the LLM response times, when combined only with the graph memory component, significantly improved when compared to ReMEmbR's vector database query times: on both \textit{Florence-2-base} and \textit{Florence-2-large} VLM configurations, the response latency over an EmbodiedLGR-Agent's LLM call decreased to half of the response times shown by the baseline. Furthermore, the global response latency when both structures were active (23.73 seconds for \textit{Florence-2-Large}) remains relatively close to the vector-database-only configuration (19.79 seconds).
Considering an equivalent average latency for each OpenAI API request across all evaluations, Table~\ref{tab:vlm_results_latency} also highlights how the response latency bottleneck is posed by the amount of LLM context consumed during the reasoning phase, as the combined agent configuration has access to a larger set of retrieval tools and tends to produce longer context windows. While latency almost doubles between Graph-only and ReMEmbR-only in both VLM configurations, it amounts to a lower $\sim$20\% latency increase between ReMEmbR-only and the combined approach (19.91\% for \textit{Florence-2-large} and 20.85\% for \textit{Florence-2-base}, respectively).
In scenarios with a prevalence of atomic queries over semantically complex ones, unlike the NaVQA dataset, the graph memory component can potentially reduce response latency under the global configuration, as it may be more heavily exploited by the architecture.

Table~\ref{tab:vlm_results_latency} also shows how often the agent invoked specific memory modules to answer queries. The LLM was prompted to rely on faster graph-querying tools whenever possible and to fall back to querying the vector database through ReMEmbR only if the information retrieved from the graph was insufficient to answer the proposed query. From the fallback percentage in the table, we observe a distinct behavioral difference across VLM implementations. The \textit{Florence-2-base} configuration exhibited a fallback percentage of 93.33\%, indicating that, for the lighter VLM implementation, the agent relied more on vector-database entries, answering 6.67\% of queries solely by accessing the graph. In contrast, \textit{Florence-2-large}, which generates semantically richer entity representations relative to the \textit{base} version, resulted in a much lower fallback percentage of 80.83\%, with 19.17\% of queries addressed solely via graph retrieval.
This shows that as the VLM-generated semantic labels become more accurate and detailed, the agent increasingly relies on the graph-memory entity representation, leading to almost 20\% of queries being answered without querying the vector database.

\section{Real-World Deployment}


To validate the lightness of the deployed EmbodiedLGR-Agent architecture, we first tested it on a traditional, consumer-grade PC. We successfully tested the ROS~2 node deployment of the agent by simulating an AgileX Scout robot in a warehouse environment with Gazebo~\cite{macenski2022ros2}, running both the memory building and query phases on an AMD Ryzen 7 3700X processor, an NVIDIA RTX3060 \SI{12}{\giga\byte} GPU, and \SI{32}{\giga\byte} of DDR4 RAM.

\begin{figure}[t]
    \centering
    \includegraphics[width=0.48\textwidth]{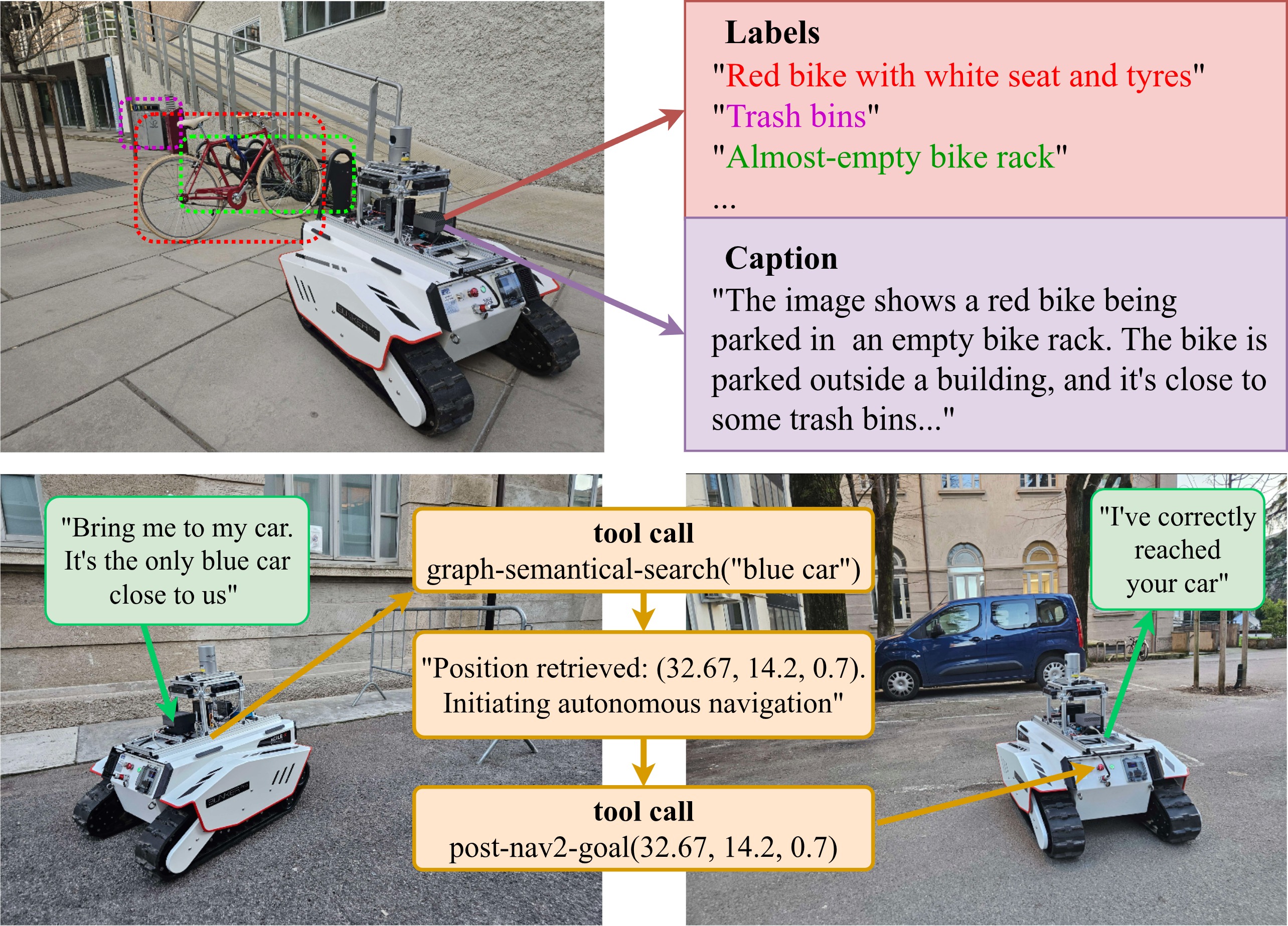}
    \caption{EmbodiedLGR-Agent was deployed on an NVIDIA Jetson Orin, interfaced with an AgileX Bunker Pro ROS~2 environment for autonomous navigation. The robot was equipped with ZED X cameras for image retrieval, and a Velodyne LiDAR for SLAM and autonomous navigation with Nav2. EmbodiedLGR-Agent learned an environment representation during the memory building phase (top) and correctly navigated to objects it had previously stored in the graph memory (bottom).}
    \label{fig:deploy}
\end{figure}

To evaluate the performance and querying capabilities of EmbodiedLGR-Agent, we then deployed it on an AgileX Bunker Pro mobile robot equipped with an NVIDIA Jetson AGX Orin \SI{64}{\giga\byte}. Our intent is to evaluate the ability of our solution to explore and build memory entries for both the memory graph and the vector database in a real-world setting with real-time processing, allowing us to test the agent on built memory at any time during and after the exploration phase. An outline of the process is shown in Fig.~\ref{fig:deploy}. 

For the employed VLM, we tested our solution with \textit{Florence-2-large} running locally on the Jetson, without any parameter quantization, given the reduced (0.77B) native model dimension. To enable integration of the ROS~2 environment~\cite{macenski2022ros2} for controlling the robot with the GPT-4o API backend, we chose \textit{ROSA} agent framework~\cite{royce2025rosa}, which is designed to allow the LLM to access the ROS~2 backend natively. 
The robot was equipped with a ZED X camera, providing an image feed over a dedicated ROS~2 topic; a Velodyne 3D LiDAR was also present on the system, allowing an estimate of the robot's pose, while also enabling autonomous navigation via Nav2~\cite{macenski2020nav2}. The agent was therefore able to access the existing ROS 2 topics on the robot's environment to retrieve the robot's pose and send autonomous navigation goals.

\textbf{EmbodiedLGR-Agent Deployment.} The EmbodiedLGR-Agent architecture was deployed over ROSA's framework as a set of ROS~2 nodes, allowing the exploitation of the ROS~2 middleware to handle data exchange between the build and querying components over the implemented memory architectures. The tools presented in Section~\ref{sec:mem_querying} were integrated directly within the LLM agent, allowing (1) on-graph semantic similarity search, (2) on-graph positional search, (3) on-graph time search, or (4) vector database retrieval through a ReMEmbR agent instance. To allow aided navigation, we also implemented a tool within the LLM, enabling the agent to forward Nav2 goal poses to the navigation stack and guide the user upon request. 

\textbf{Results.} An overview of the deployed system, performing both memory building and agent query tasks, is shown in Fig.~\ref{fig:deploy}. The agent is evaluated qualitatively by selecting specific entities and events observed during exploration to be queried on, in order to evaluate the practical utility of the architecture in a real-world setting. We first drove the robot around a university campus for 30 minutes, running the memory-building pipeline as it explored portions of the space. We then queried the EmbodiedLGR-Agent with semantic, positional, or temporal requests to stress-test specific tool invocations and to evaluate the agent's ability to suggest specific locations from generic questions. For example, we provided questions such as ``Take me to a place where I can repair my robot" to which the robot correctly replied ``I found the location of a `laboratory' at position $\{x,y,z\}$. Follow me", having observed the word ``laboratory" on the corresponding building's entrance door during the exploration phase. 

Other queries, such as ``Find me a bench with a red wooden tint", were not addressed correctly as the robot returned a set of known benches in memory. This is likely due to the way the environment was explored, as it was conducted while the robot was remotely controlled at a maximum speed of 1.5 meters per second. For this reason, brief observations of objects during rapid motion were not processed into semantically detailed labels by the VLM. 

Apart from specific, semantically complex queries about object details, the robot correctly returned navigable positions from memory and exhibited human-like responsiveness across diverse queries. Human-robot interaction was particularly responsive for simple, atomic queries like ``where is the closest hydrant?", fully exploiting the memory graph query tools, where the answering overhead was basically limited only by the LLM's APIs inference times.

\section{Conclusions}

In this work, we introduced EmbodiedLGR-Agent, a novel VLM-driven agent architecture designed to build and retrieve efficient semantic-spatial memories for robotic agents in real-time scenarios. Our agent achieves highly responsive inference times suitable for real-world human-robot interaction while allowing for exploration and the direct construction of memory representations within the robot's deployment window. The proposed EmbodiedLGR-Agent enables robots to leverage long-term memory, presenting a novel state-of-the-art approach to efficient and scalable memory management for intelligent agents operating under time-critical or computationally bounded conditions. 
The memory building and retrieval pipelines, although constrained in semantic density by the choice of a reduced-size family of models like \textit{Florence-2}, are evaluated on the NaVQA dataset, proving competitive accuracy within region-level spatial and temporal boundaries in an edge-deploy condition.

\textbf{Limitations and Future Work.}
The proposed architecture presents some limitations. It depends heavily on the underlying LLM, which can occasionally produce incorrect tool calls or misinterpret retrieved information. We also deliberately use small, multi-purpose VLMs such as \textit{Florence-2} to enable real-time, low-latency deployment, but this reduces the richness and granularity of scene descriptions relative to larger models. Future work will evaluate alternative LLMs for more reliable tool invocation and explore other edge-deployable VLMs. 
Moreover, as the current spatial representation from both memory structures is referenced from the robot's pose reference system, future work could address 3D spatial grounding of semantic entities to evaluate the correctness of the environment reconstruction against the ground truth.
Finally, beyond our efficient memory structure, we plan to investigate hierarchical memory representations to support spatial reasoning over complex yet efficiently traversable memory-graph structures for real-time settings.

\bibliographystyle{IEEEtran}
\bibliography{IEEEabrv, mybib}

\end{document}